%% file: EviCare.tex
\documentclass[sigconf]{acmart}

\usepackage{algorithmic}
\usepackage{graphicx}
\usepackage{multirow}
\usepackage{amsmath,amssymb,amsfonts}
\usepackage{subfig}
\usepackage{arydshln}
\usepackage[shortlabels]{enumitem} 
\usepackage{appendix} 
\usepackage{chngcntr} 
\usepackage{xcolor}
\usepackage[table]{xcolor}
\usepackage{booktabs}
\usepackage{xspace}
\usepackage{balance}
\usepackage{tcolorbox}
\usepackage{array}
\definecolor{blueblack}{HTML}{000080}

\newcommand{\blue}[1]{\textcolor{blue}{#1}}

\usepackage[linesnumbered,vlined,ruled]{algorithm2e}
\newcommand{\method}{\texttt{EviCare}\xspace}

\AtBeginDocument{%
  }

\copyrightyear{2026}
\acmYear{2026}
\setcopyright{cc}
\setcctype{by}
\acmConference[KDD '26]{Proceedings of the 32nd ACM SIGKDD Conference on Knowledge Discovery and Data Mining V.1}{August 09--13, 2026}{Jeju Island, Republic of Korea}
\acmBooktitle{Proceedings of the 32nd ACM SIGKDD Conference on Knowledge Discovery and Data Mining V.1 (KDD '26), August 09--13, 2026, Jeju Island, Republic of Korea}
\acmPrice{}
\acmDOI{10.1145/3770854.3780257}
\acmISBN{979-8-4007-2258-5/2026/08}

\begin{document}

\title{\method: Enhancing Diagnosis Prediction with Deep Model-Guided Evidence for In-Context Reasoning}

\author{Hengyu Zhang}

\email{hengyu.zhang3@hdr.mq.edu.au}
\affiliation{%
  \institution{Macquarie University}
  \city{Sydney}
  \country{Australia}
}

\author{Xuyun Zhang}
\email{xuyun.zhang@mq.edu.au}
\affiliation{%
  \institution{Macquarie University}
  \city{Sydney}
  \country{Australia}
  }

\authornote{Xuyun Zhang is the corresponding author.}

\author{Pengxiang Zhan}
\email{yyytdms@gmail.com}
\affiliation{%
  \institution{Fuzhou University}
  \city{Fuzhou}
  \country{China}
}

\author{Linhao Luo}
\email{Linhao.Luo@monash.edu}
\affiliation{%
 \institution{Monash University}
 \city{Melbourne}
 \country{Australia}
 }

\author{Hang Lv}
\email{lvhangkenn@gmail.com}
\affiliation{%
  \institution{Fuzhou University}
  \city{Fuzhou}
  \country{China}
  }

\author{Yanchao Tan}
\email{yctan@fzu.edu.com}
\affiliation{%
  \institution{Fuzhou University}
  \city{Fuzhou}
  \country{China}
  }

\author{Shirui Pan}
\email{s.pan@griffith.edu.au}
\affiliation{%
  \institution{Griffith University}
  \city{Brisbane}
  \country{Australia}
  }

\author{Carl Yang}
\email{j.carlyang@emory.edu}
\affiliation{%
  \institution{Emory University}
  \city{Atlanta}
  \country{USA}
  }

\renewcommand{\shortauthors}{Hengyu Zhang et al.}

\begin{abstract}

Recent advances in large language models (LLMs) have enabled promising progress in diagnosis prediction from electronic health records (EHRs). However, existing LLM-based approaches tend to overfit to historically observed diagnoses, often overlooking novel yet clinically important conditions that are critical for early intervention. To address this, we propose \method, an in-context reasoning framework that integrates deep model guidance into LLM-based diagnosis prediction. Rather than prompting LLMs directly with raw EHR inputs, \method performs (1) deep model inference for candidate selection, (2) evidential prioritization for set-based EHRs, and (3) relational evidence construction for novel diagnosis prediction.
These signals are then composed into an adaptive in-context prompt to guide LLM reasoning in an accurate and interpretable manner.
Extensive experiments on two real-world EHR benchmarks (MIMIC-III and MIMIC-IV) demonstrate that \method~\footnote{https://github.com/zhyccc/EviCare} achieves significant performance gains, which consistently outperforms both LLM-only and deep model-only baselines by an average of 20.65\% across precision and accuracy metrics.
The improvements are particularly notable in challenging novel diagnosis prediction, yielding average improvements of 30.97\%.

\end{abstract}





\maketitle

\input{chapter/introduction.tex}

\input{chapter/related.tex}
\input{chapter/method}

\input{chapter/experiment}

\input{chapter/conclusion}

\section*{Acknowledgements}
{This research was supported by the Commonwealth through an Australian Government Research Training Program Scholarship, the State Key Laboratory of Novel Software Technology (KFKT2024A03), the Fujian Provincial Artificial Intelligence Industry Development Technology Project under Grant (2025H0042), and Fujian Provincial Natural Science Foundation of China under Grants (2025J01540).} 
Carl Yang was not supported by any funds from China. 

\newpage
\bibliographystyle{ACM-Reference-Format}
\bibliography{EviCare}
\balance

\newpage

\appendix
\input{chapter/appendix}
\end{document}

%% file: chapter/introduction.tex
\begin{figure}
    \centering
    \includegraphics[width=\linewidth]{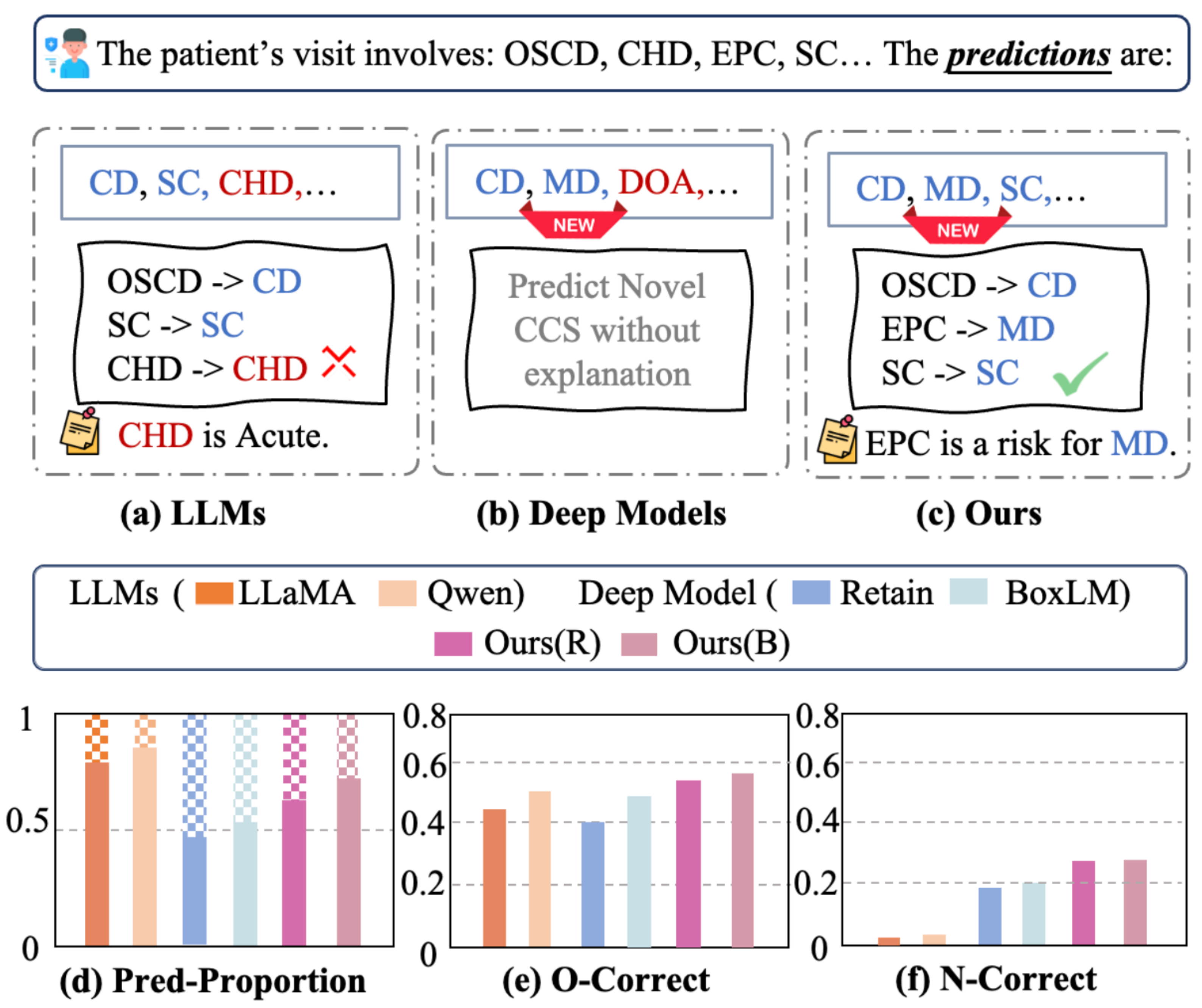}
    \caption{Diagnosis Prediction Patterns Across Models.
(a) LLMs tend to predict repeating historical diagnoses (e.g., CHD);
(b) Deep models are effective at predicting novel diagnoses (e.g., MD) but lack explanation;
(c) Our method integrates deep model-guided evidence into LLMs to achieve accurate and explainable predictions (e.g., {EPC} $\rightarrow$ {MD}).
(d) Proportion of predictions: historical (solid bars) vs. novel (hatched bars);
(e) Overall prediction performance (Precision@10, O-Correct);
(f) Performance on novel diagnoses (Precision@10, N-Correct).}
    \label{fig:toy_example}
\end{figure}

\section{Introduction}
\label{sec:intro}

Accurate diagnosis prediction from electronic health records (EHRs) plays a vital role in clinical decision-making, enabling early detection of complications and personalized treatment planning~\cite{chiu2023integrating, antikainen2023transformers}. With growing access to longitudinal EHR data, recent studies have begun exploring large language models (LLMs) for diagnosis prediction~\cite{chen2024rarebench, clinicalcot2024, dearllm2025, medagents2024}. These approaches leverage LLMs' biomedical knowledge and reasoning capabilities to support medical decision-making. For example, Kwon et al.~\cite{clinicalcot2024} propose a reasoning-aware prompting strategy to generate interpretable diagnostic rationales. DearLLM~\cite{dearllm2025} enhances diagnosis prediction by extracting feature correlations deduced by LLMs to support personalized care.

Despite promising results for LLM-based diagnosis prediction, a critical unmet challenge remains: accurately predicting novel diagnoses, namely those not previously recorded in a patient’s history but clinically relevant for proactive care. These diagnoses often indicate disease progression, new comorbidities, or acute deterioration, making them highly valuable for early intervention. However, the task requires temporal reasoning over patient-specific visit sequences that vary widely across individuals.

As illustrated in Figure~\ref{fig:toy_example}(a), LLMs often repeat previously observed diagnoses. For instance, predicting {CHD} (Coagulation and Hemorrhagic Disorders) merely because it appeared in earlier visits, even though it is typically an acute condition unlikely to persist. This behavior arises not from hallucination, but from the LLM’s next-token prediction objective~\cite{radford2019language}, which favors contextually familiar outputs. While LLMs generate a mix of historical and novel predictions (Figure~\ref{fig:toy_example}(d)), the majority are historical. The further demo experiment shows that they are accurate primarily on those repeated diagnoses (Figure~\ref{fig:toy_example}(e)), and fail on novel ones (Figure~\ref{fig:toy_example}(f)), exposing a clear limitation in discovery-oriented reasoning.

In contrast, deep models (e.g., Retain~\cite{choi2016retain} and BoxLM~\cite{tan2025boxlm}) trained on EHR data better capture temporal and structured patterns, enabling detection of previously unseen conditions and outperforming LLMs on novel diagnosis prediction. As shown in Figure~\ref{fig:toy_example}(b), these models successfully surface {MD} (Mood Disorders) and {DOA} (Deficiency and Other Anemia) given relevant comorbid signals like {EPC} (Epilepsy and Convulsions). However, these deep models are fundamentally restricted to categorical outputs. While they can effectively identify what the likely diagnosis is based on latent patterns, they lack the generative capability to articulate the clinical reasoning process (e.g, explaining why lead to the prediction), which is essential for verifiable clinical decision-making.

To address these complementary limitations, we propose \method, a hybrid reasoning framework that transforms numerical signals from deep models into structured clinical reasoning. Rather than relying solely on LLMs to reason over raw EHR inputs, we extract structured evidence from a deep model and embed it into an adaptive in-context prompt to guide LLM reasoning.

Specifically, \method incorporates four core evidential components:
(1) {Deep model inference for candidate selection}, which identifies top-$K$ diagnoses covering both historical and novel using prediction logits from a deep model;
(2) {Evidential prioritization for set-based EHRs}, which ranks historical diagnoses based on their contribution to the deep model's prediction, transforming unordered EHR codes into weighted organized inputs;
(3) {Relational evidence construction for novel prediction}, which builds symbolic links between historical and candidate diagnoses by extracting co-occurrence and ontology-based relations from large-scale EHRs, helping justify novel predictions;
(4) {LLM-based diagnosis via evidence-driven in-context learning},  which composes the above components into a structured prompt that guides the LLM to perform accurate and interpretable predictions.
As shown in Figure~\ref{fig:toy_example}(c), (e), and (f), \method improves both novel and overall diagnosis accuracy, and provides interpretable reasoning.

We evaluate \method on two real-world benchmarks (MIMIC-III and MIMIC-IV) and under both overall and novel prediction settings. Experiments demonstrate that \method consistently outperforms LLM-only and deep-only baselines, particularly in challenging novel diagnosis scenarios, while offering transparent and evidence-grounded outputs.

Our contributions are summarized as follows:
\begin{itemize}[leftmargin=10pt]
\item \textit{Identification of LLM Limitations in Diagnosis Prediction.} We reveal key limitations of LLMs in EHR-based diagnosis, including a tendency toward repeating historical conditions and poor performance on clinically important novel diagnoses.

\item \textit{A Deep Model-Guided Reasoning Paradigm.} We introduce \method, a novel hybrid reasoning framework that bridges deep EHR models with LLMs via structured evidence-guided prompting. This paradigm enables LLMs to reason beyond surface patterns by leveraging deep model-generated signals through three aligned mechanisms: Candidate Selection, Evidential Prioritization, and Relational Evidence Construction.

\item \textit{Comprehensive Evaluation on Real-World EHRs.} We validate the proposed \method on two large-scale clinical datasets (MIMIC-III and MIMIC-IV), showing consistent improvements over LLM-only and deep-only baselines. Our approach significantly enhances novel diagnosis prediction while maintaining interpretability and generalizability under limited supervision.
\end{itemize}

%% file: chapter/related.tex
\section{Related Work}

\subsection{Diagnosis Prediction with Deep Models}

Diagnosis prediction from electronic health records (EHRs) has become a central problem in clinical informatics, essential for early detection and personalized care~\cite{hendriksen2013diagnostic, nasir2024ethical,jiu2024literature,aggarwal2025harnessing}. Existing deep learning approaches can be broadly categorized into two categories: 

The first line of work focuses on capturing temporal and contextual dependencies across patient visits using neural sequence models. For instance, RETAIN~\cite{choi2016retain} and Dipole~\cite{ma2017dipole} encode visit sequences to learn time-aware patient embeddings, while Transformer-based methods~\cite{chen2024predictive} enhance the modeling of long-range dependencies and clinical state transitions via self-attention mechanisms.

The second line of work augments patient representations using structured medical knowledge. 
For example, GRAM~\cite{choi2017gram} incorporated hierarchical relations from clinical ontologies via an attention mechanism. CGL~\cite{lu2021collaborative} jointly modeled ontology and co-occurrence signals through graph learning. SeqCare~\cite{xu2023seqcare} leveraged personalized knowledge graphs and label dependencies to suppress task-irrelevant noise. More recently, geometry-inspired models such as BoxCare~\cite{lv2024boxcare} and BoxLM~\cite{tan2025boxlm} encoded medical concepts as high-dimensional boxes to capture inclusion hierarchies and semantic proximity.
Despite their accuracy, these models operated in latent spaces without offering explicit reasoning, hindering their integration with LLMs for interpretable clinical decision-making.

\begin{figure*}
    \centering
    \includegraphics[width=1\linewidth]{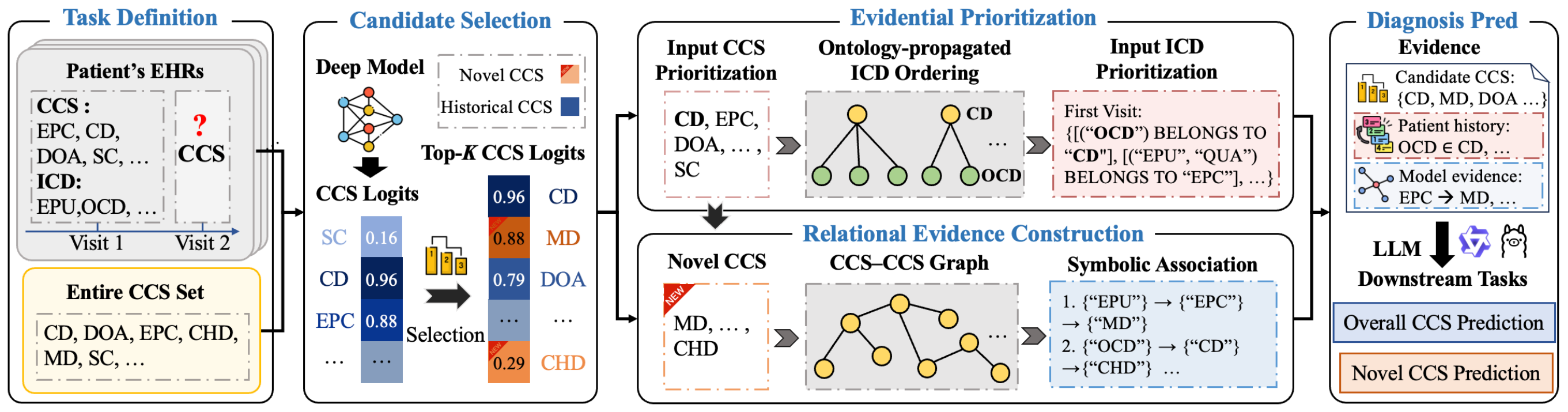}
    \caption{The overall framework of our proposed \method, which consists of (1) deep model inference for \underline{Candidate Selection}, (2) \underline{Evidential Prioritization} for set-based EHRs, (3) \underline{Relational Evidence Construction} for novel diagnosis prediction, and (4) LLM-based \underline{Diagnosis Pred}iction via evidence-driven in-context learning.}
    \label{fig:Framework}
\end{figure*}

\subsection{Large Language Models for Healthcare}

Recent years have witnessed the rapid progress of large language models (LLMs) in the medical domain~\cite{nori2023capabilities, kasai2023evaluating}. Foundation models like GPT-4 have demonstrated impressive performance in medical question answering, radiology interpretation, and differential diagnosis~\cite{jin2024genegpt,zheng2025comparison}, with some specialized LLMs (e.g., AMIE~\cite{tu2025towards}) even surpassing the diagnostic accuracy of primary care physicians. These models not only extract clinical evidence from text but also support clinical reasoning by generating fluent and coherent diagnostic rationales~\cite{savage2024diagnostic}.

To enhance LLMs' clinical utility, two major research directions have emerged. The first is tool-augmented prompting, where external medical knowledge or similar patient records are retrieved and provided as context for LLMs~\cite{jin2024health}. In this way, LLMs are empowered by retrieval~\cite{zakka2024almanac}, search APIs~\cite{jin2024genegpt}, or agent-based systems~\cite{wu2024ehrflow} to incorporate up-to-date clinical knowledge. Recent work such as KARE~\cite{jiang2024reasoning} further extends this line by organizing large-scale medical knowledge graphs into community-level structures, enabling more focused retrieval to support LLM-based healthcare prediction and mitigate hallucinations. The second focuses on instruction tuning, where LLMs are adapted to the medical domain using collected datasets from biomedical literature, knowledge bases, or self-generated rationales~\cite{ wang2024survey}. These approaches significantly improve LLMs’ generalization in medical tasks, including diagnosis, prognosis, and treatment planning.

However, existing LLM-based methods often overlook the temporal structure and symbolic relationships embedded in EHRs. While recent approaches like DearLLM~\cite{dearllm2025} and CKLE~\cite{ding2024distilling} have attempted to incorporate code frequency, co-occurrence patterns, and ontology-based signals into LLM reasoning, they still fall short in effectively capturing patient-specific relational context. Related efforts also explore integrating structured knowledge with EHRs, such as GraphCare~\cite{jiang2023graphcare}, which constructs personalized patient knowledge graphs using LLMs and external biomedical KGs for downstream prediction, and InKrat~\cite{li2025inkrat}, which leverages cross-modal semantic retrieval and LLM-generated explanations to enhance interpretability. As a result, LLMs tend to treat diagnosis prediction as a sequence generation task, leading to over-repetition of historical codes and limited recognition of novel conditions. These limitations underscore the need for a framework that grounds LLM reasoning in structured, clinically meaningful evidence derived from the EHR itself.


%% file: chapter/method.tex
\section{THE \method FRAMEWORK}
\label{sec:method}
\subsection{Problem Setup and Framework Overview}
We consider the task of diagnosis prediction from longitudinal electronic health records (EHRs). Each patient $p$ is represented by a sequence of historical visits $\mathcal{V}_p = \{v_1, v_2, \dots, v_t\}$, Where each visit $v_i$ is associated with a set of diagnosis codes drawn from either the ICD (\emph{International Classification of Diseases}) or CCS (\emph{Clinical Classifications Software}) vocabulary.
Since ICD represents specific diagnoses, while CCS further groups ICD codes, we involve both of them to encode hierarchical and complementary semantics.

Given this visit history $\mathcal{V}_p$, the goal is to predict the relevant CCS codes $\hat{\mathcal{C}}_{t+1}$ for the patient’s next visit $v_{t+1}$. We formulate two sub-tasks: 

(1) \textbf{Overall diagnosis prediction}, which includes all clinically relevant CCS codes regardless of whether they have appeared in history;
(2) \textbf{Novel diagnosis prediction}, which filters out previously observed diagnoses to assess the model's capacity for discovering unseen yet clinically consistent outcomes.
Both tasks share the same EHR input, but differ in label construction.

We summarize the main modules of the \method framework in Figure~\ref{fig:Framework} to provide an overview. The first component, \textit{Deep Model-Guided Candidate Selection}, uses a deep model (e.g., BoxLM or RETAIN) to identify the top-$K$ most probable diagnosis codes, grounding the LLM’s reasoning within a clinically relevant candidate space.
The second component, \textit{Historical Evidence Prioritization}, ranks historical diagnoses based on their predictive contribution as determined by the deep model, transforming unordered EHR codes into a weighted and structured representation.
The third component, \textit{Relational Evidence Construction}, builds symbolic links between historical and candidate diagnoses using co-occurrence statistics and ontology hierarchies, offering interpretable justification for novel predictions.
The fourth component, \textit{LLM-based Diagnosis via Evidence-driven In-Context Learning}, composes all evidence into a structured prompt that guides the LLM toward accurate and explainable diagnostic reasoning.

\subsection{Deep Model Inference for Candidate Selection}
As shown in Figure~\ref{fig:toy_example}, while LLMs exhibit strong reasoning capabilities, they often struggle to make clinically grounded predictions when faced with structured EHR inputs, frequently overemphasizing previously seen diagnoses. In contrast, deep models trained on EHR data provide a more reliable estimation of diagnosis probabilities by learning task-specific relational patterns. However, such models typically lack interpretability.

To combine the strengths of both paradigms, we propose to leverage deep model prediction logits as a compact and model-agnostic representation of diagnosis likelihood. These logits enable us to construct a candidate diagnosis set $\mathcal{C}_p^{\text{cand}}$ by selecting the top-K most probable CCS codes for each patient visit. Rather than being a heuristic filter, this design is theoretically motivated by the principle that the ranked logits list from deep models encodes a structured approximation of the underlying information distribution. Let $y_{p, c}$ be the deep model logit for patient $p$ and candidate diagnosis $c$. We formalize the theoretical guarantees as follows:

\begin{lemma}[Ranked-Logit Information Prior and Denoising]
The ranked-logit prior establishes the following properties:
\begin{enumerate}
    \item Conditioning on $\mathcal{C}_p^{\text{cand}}$ reduces the conditional entropy $H(\mathcal{C}\mid X_p)$ and increases mutual information, thereby effectively denoising the reasoning space;
    \item The Top-$K$ selection by posterior probability is Bayes-optimal under a $K$-budget constraint~\cite{gupta2025enough};
    \item Using rank-based weights as priors does not increase Bayes risk compared to flat prompting.
\end{enumerate}
\end{lemma}

This analysis provides a theoretical justification for using ranked logits as an information prior, explaining why constraining the candidate space improves robustness without sacrificing flexibility.

In practive, this design grounds the LLM’s inference within a clinically relevant scope while avoiding open-ended generation, which often leads to noisy or redundant outputs. Importantly, using logits ensures compatibility across different deep architectures. By extracting only the final probabilistic output, our framework remains lightweight and flexible, supporting plug-and-play integration with diverse backbone models.

We instantiate this component with two representative deep architectures (BoxLM~\cite{tan2025boxlm} due to its effective performance and RETAIN~\cite{choi2016retain} due to its wide adoption) to illustrate the framework’s generality, where both output clinically grounded prediction logits can be directly used in downstream reasoning.

\noindent\textbf{BoxLM (Structure-Aware Logits).} BoxLM encodes each diagnosis (ICD or CCS) as a high-dimensional box \( \mathbf{b}_i = (\mathbf{b}^{\text{Cen}}_i,\ \mathbf{b}^{\text{Off}}_i) \), where the center $\mathbf{b}^{\text{Cen}}_i$ encodes the semantic core (initialized from BioBERT) and the offset $\mathbf{b}^{\text{Off}}_i$ represents conceptual uncertainty (learned via ontology-aware GCN). Visit-level and patient-level representations are aggregated via attention and temporal pooling:
\begin{equation}
\mathbf{b}_{v_t}^{\text{Cen}} = \sum_{i \in v_t} \alpha_i \mathbf{b}_i^{\text{Cen}}, \quad
\mathbf{b}_{v_t}^{\text{Off}} = \max_{i \in v_t} \mathbf{b}_i^{\text{Off}},
\end{equation}
\begin{equation}
\mathbf{b}_p^{\text{Cen}} = \sum_{v_i \in \mathcal{V}_p} w_i \mathbf{b}_{v_i}^{\text{Cen}}, \quad
\mathbf{b}_p^{\text{Off}} = \max_{v_i \in \mathcal{V}_p} \mathbf{b}_{v_i}^{\text{Off}},
\end{equation}
where $i$ indexes diagnosis codes within visit $v_t$, and $v_t$ denotes the $t$-th visit of patient $p$.

We calculate the relevance of a candidate CCS \( c \) by measuring the intersection volume between patient and candidate boxes:
\begin{equation}
\hat{y}_{p,c} = \log\left(\text{Vol}(\mathbf{b}_p \cap \mathbf{b}_c)\right),
\end{equation}
where the intersection volume is computed using a differentiable approximation. Specifically, we adopt a \emph{Gumbel-softplus approximation} to estimate the volume of intersection between two boxes in a smooth and numerically stable manner:
\begin{equation}
\text{Vol}(\mathbf{b}_p \cap \mathbf{b}_c) \approx \prod_{k=1}^{d} \beta \log\left(1 + \exp\left(-\frac{\mu^{\text{max}}_k - \mu^{\text{min}}_k}{\beta} - 2\gamma \right)\right),
\end{equation}
where $\mu_k^{\text{min}}$ and $\mu_k^{\text{max}}$ are minimum and maximum corners corresponding to the intersection box between $\mathbf{b}_p$ and $\mathbf{b}_c$. $\beta$ is the scale of the Gumbel distribution and $\gamma$ is the Euler-Mascheroni constant.

\noindent\textbf{RETAIN (Sequence-Aware Logits).} RETAIN computes diagnosis prediction scores by attending to historical visit representations in reverse chronological order. Each visit $x_t \in \mathbb{R}^r$ is embedded as $
\mathbf{v}_t$. Two RNNs are applied over the reversed visit sequence $\{\mathbf{v}_T, \ldots, \mathbf{v}_1\}$ to generate visit-level attention weights $\alpha_t$ and variable-level attention vectors $\beta_t$.

Then, the patient representation is aggregated via:
\begin{equation}
\mathbf{c}_p = \sum_{t=1}^{T} \alpha_t \cdot (\beta_t \odot \mathbf{v}_t).
\end{equation}

The prediction logits for each candidate CCS are computed via:
\begin{equation}
\hat{y}_{p,c} = \left( W_o \cdot \mathbf{c}_p + b_o \right)_c.
\end{equation}

\noindent\textbf{Unified Candidate Selection.}
Regardless of architecture, we extract the top-$K$ CCS codes based on logits $\hat{y}_{p,c}$, and define these as the candidate set $\mathcal{C}_p^{\text{cand}}$ for downstream LLM-based reasoning. This modular design ensures that any diagnosis model producing logits—regardless of its internal structure—can be integrated into our framework, facilitating broad applicability across settings.

\subsection{Evidential Prioritization for Set-based EHRs}

Raw EHR inputs are typically represented as unordered sets of ICD/CCS codes, lacking clinical prioritization or semantic hierarchy. However, not all diagnoses contribute equally to a patient’s current state. When such flat, unstructured code sets are provided to LLMs, their ability to focus on the most relevant clinical cues is diminished, leading to noisy or generic reasoning.

To address this, we introduce an evidential prioritization strategy that enhances both the informativeness and structure of the patient history used in prompting. Specifically, we first rank each historical CCS code by computing its marginal contribution to the deep model’s prediction logits. Then, using the CCS–ICD ontology\footnote{https://www.hcup-us.ahrq.gov/toolssoftware/ccs/ccs.jsp}, we propagate these priorities to all corresponding ICD codes. The intuition is to quantify each diagnosis’s contribution to the deep model’s prediction and reflect that in the prompt structure.

\noindent\textbf{Input CCS Prioritization.}
To highlight clinically relevant conditions within the unordered historical CCS codes $\mathcal{H}_p$, we rank them by their predictive influence using the deep model’s output logits. This prioritization helps guide the LLM to focus on diagnoses that most impact the model's predictions.
This choice is both model-agnostic and task-aligned: unlike attention weights or hidden states that vary significantly across architectures, logits provide a consistent and interpretable signal that reflects each input code’s contribution to the model’s prediction. Clinically, this prioritization helps emphasize persistent or influential conditions (e.g., chronic comorbidities or early indicators of disease onset), which are known to shape diagnostic outcomes over time~\cite{choi2016retain, ma2017dipole}.  Then, we convert the unordered set $\mathcal{H}_p$ into an ordered list $[\tilde{C}_1, \tilde{C}_2, \ldots]$ by sorting according to:
\begin{equation}
[\tilde{C}_1, \tilde{C}_2, \ldots] = \text{Sort}_{h \in \mathcal{H}_p}(\hat{y}_{p,h}).
\end{equation}
This ordering reflects the model’s estimation of how well each CCS code aligns with the patient’s current condition, enabling downstream prompt construction to emphasize more relevant diagnoses.

\noindent\textbf{Ontology-Propagated ICD Ordering.}
In most diagnostic models, prediction is performed on either CCS or ICD code space—but not both. As a result, we typically obtain prediction logits only for one level of the ICD-CCS ontology, while the other lacks direct supervision. In our setting, the deep model outputs logits for CCS codes, leaving the ICD-level relevance unspecified.

To incorporate fine-grained ICD-level detail while maintaining the diagnostic relevance provided by CCS-level prediction, we propagate CCS priorities to their associated ICD codes based on a standardized ontology.

Formally, for each CCS code $C_h$ with prediction score $\hat{y}_{p,h}$, we collect the corresponding ICD codes from the historical visit as:
\begin{equation}
\mathcal{S}_h = \{ \text{ICD}_i \in \mathcal{V}_p \mid \text{ICD}_i \text{ belongs to } C_h \},
\end{equation}
where ``belongs to" follows the CCS–ICD ontology mapping.

Each ICD group $\mathcal{S}_h$ inherits the priority of its parent CCS $C_h$, allowing us to present historical diagnoses in a relevance-aware, semantically structured format. A corresponding prompt segment is constructed as:
\begin{tcolorbox}[colback=blueblack!8!white, 
colframe=blueblack!85!black, 
arc=2mm, auto outer arc,
title=Evidential Prioritization for Set-based EHRs]
\textbf{Patient Historical Diagnoses (Prioritized):}  

\texttt{[\{"Essential Hypertension", "Hypertensive Heart Disease"\} BELONG TO "Hypertension"],} 
\texttt{[\{"Type 2 Diabetes without Complications"\} BELONG TO "Diabetes Mellitus"], ...]}

\end{tcolorbox}
This formulation enhances the prompt with both concept-level relevance and fine-grained clinical specificity, enabling the LLM to reason over patient history in a hierarchically informed and interpretable manner.

\subsection{Relational Evidence Construction for Novel Diagnosis Prediction}

As shown in Figure~\ref{fig:toy_example}, LLMs often struggle to infer novel diagnoses absent from a patient's history due to the lack of explicit supporting evidence. Without structured cues, they tend to repeat previously seen conditions or overlook emerging comorbidities. In contrast, deep models trained on large-scale EHR datasets can implicitly capture rich statistical relationships between diagnosis codes, enabling them to assign high scores even to novel diagnoses through learned co-occurrence patterns.

However, such reasoning remains hidden within the deep model's black-box architecture. To enhance interpretability and support LLM-based inference, we extract symbolic relational evidence between historical and candidate CCS codes. Specifically, we mine co-occurrence statistics from the training data to discover potential associative links between past diagnoses and each top-ranked novel candidate identified by the deep model. 
While the construction process itself is model-agnostic, we selectively extract symbolic relations only for the novel candidate diagnoses predicted by the deep model.
By bridging global co-occurrence with instance-specific deep model predictions, we enable the LLM to reference relevant medical associations that may reflect latent patterns implicitly captured by the deep model.
Specifically, we extract such relational evidence in two steps: (1) constructing a CCS–CCS co-occurrence matrix from patient-level diagnosis records, and (2) identifying, for novel candidate CCS generated by deep models, the most statistically related historical CCS from the patient’s record.

\noindent\textbf{CCS Co-occurrence Matrix Construction.}
Let $\mathcal{A} \in \mathbb{R}^{N \times C}$ denote the patient-to-CCS binary adjacency matrix, where $N$ is the number of patients and $C$ is the number of CCS codes. Each entry $\mathcal{A}(n, c) = 1$ indicates that patient $n$ was diagnosed with CCS code $c$. The CCS co-occurrence matrix is then defined as:
\begin{equation}
\mathbf{G}_{\text{ccs}} = \mathcal{A}^\top \cdot \mathcal{A},
\end{equation}
where $\mathbf{G}_{\text{ccs}}(i, j)$ denotes the total number of patients who have been diagnosed with both $C_i$ and $C_j$. This matrix encodes diagnosis associations that are implicitly used by deep models during learning.

\noindent\textbf{Historical-to-Candidate Relation Extraction.}
Let $\mathcal{C}_p^{\text{cand}}$ be the set of deep model-generated candidate CCS codes for patient $p$, and $\mathcal{H}_p$ the set of their historical CCS codes. For each candidate $C_c \in \mathcal{C}_p^{\text{cand}} \setminus \mathcal{H}_p$ that is not previously observed, we identify the most related historical diagnosis $C_h^*$ using the co-occurrence matrix:
\begin{equation}
C_h^* = \arg\max_{C_h \in \mathcal{H}_p} \mathbf{G}_{\text{ccs}}(C_h, C_c).
\end{equation}
This relation $C_h^* \Rightarrow C_c$ suggests that the novel diagnosis $C_c$ is statistically supported by its co-occurrence with $C_h^*$ in historical patient records. It can be interpreted as an externalized rationale for why the deep model considers $C_c$ clinically plausible, and subsequently passed to the LLM to enhance its reasoning process.

\begin{tcolorbox}[colback=blueblack!8!white, 
colframe=blueblack!85!black, 
arc=2mm, auto outer arc,
title=Relational Evidence for Novel Diagnosis Inference]
\textbf{Relational Evidence Support:}

\texttt{"Cardiac dysrhythmias"} $\Rightarrow$ \texttt{"Conduction disorders"} \\
\texttt{"Hypertension"} $\Rightarrow$ \texttt{"Pulmonary heart disease"} ...
\end{tcolorbox}
These historical-to-candidate relations offer interpretable, model-aligned support for novel predictions. By explicitly encoding statistical justifications into the LLM prompt, we enable the model to go beyond memorized history and generalize toward clinically coherent yet previously unobserved diagnoses.

\subsection{LLM-based Diagnosis Prediction via Evidence-driven In-Context Learning}

Building on the structured evidential signals derived from deep model outputs, our framework enables large language models (LLMs) to perform clinical diagnosis prediction through evidence-guided in-context reasoning. 
Instead of relying on model fine-tuning, we inject task-relevant information, including filtered candidate sets, prioritized historical diagnoses, and relational associations to novel conditions, into the prompt in a structured and interpretable format.

This framework accommodates both \emph{overall diagnosis prediction}, where the LLM observes all historical visits and may output both recurring and new conditions, and \emph{novel diagnosis prediction}, where only the most recent visit is retained and previously seen diagnoses are excluded from the candidate list. In both settings, the prompt is composed from three evidence types extracted by prior modules: prioritized diagnosis history, relational support for novel conditions, and a task-specific candidate list.

An example of the structured prompt for the \emph{novel diagnosis prediction} task is shown below:

\begin{tcolorbox}[colback=blueblack!8!white, 
colframe=blueblack!85!black, 
arc=2mm, auto outer arc,
title=Prompt for Novel Diagnosis Prediction]
\textbf{Last Diagnostic Visit (8 days ago):} [\ldots]

\textbf{Evidential Prioritization for Set-based EHRs:} [\ldots]

\textbf{Relational Evidence for Novel Diagnoses:} [\ldots]

\textbf{Candidate CCS Codes (Novel Only):} [\ldots]

\textbf{Instruction:}  
- Re-rank the candidate CCS categories from most to least likely.\\
- Output format: \texttt{Answer: <CCS 1>, <CCS 2>, ...}
\end{tcolorbox}

This unified prompting strategy allows LLMs to flexibly perform both general and novel diagnosis prediction in an interpretable, evidence-grounded manner—without any additional model training or adaptation.

%% file: chapter/experiment.tex
\section{Experiments}
\label{sec:exp}

We conduct comprehensive experiments to evaluate  the predictive performance, reasoning ability, and generalization capability of our proposed \method framework. Our evaluation seeks to answer the following research questions:

\begin{itemize}[leftmargin=10pt]
  \item \textbf{RQ1:} How does \method perform compared with existing state-of-the-art methods in diagnosis prediction?
  
  \item \textbf{RQ2:} What is the contribution of each component in \method to overall and novel diagnosis prediction?

  \item \textbf{RQ3:} How does \method compare to existing reasoning approaches in predicting novel diagnoses?

  \item \textbf{RQ4:} How do factors such as candidate size and LLM backbone affect prediction performance?
  
  \item \textbf{RQ5:} Does \method provide clinically valid and interpretable justifications for its predictions?
\end{itemize}

\begin{table}[]
    \centering
    \caption{Statistics of the datasets used in our experiments.}
    \begin{tabular}{l|cc}
    \toprule
    Dataset & MIMIC-III & MIMIC-IV \\
    \hline
    \# of patients                  & 5,449         & 79,393 \\
    \# of visits                    & 14,141        & 329,605 \\
    Avg. \# visits per patient      & 2.60          & 4.15 \\
    Avg. \# CCS per visit           & 12.08         & 11.30 \\
    Avg. \# Novel CCS per visit     & 6.15          & 4.82 \\
    Max. \# visits per patient      & 29            & 169 \\
    \hline
    \# of unique diagnoses          & 3,874          & 37,917 \\
    \# of CCS codes                 & 285           & 842 \\
    \bottomrule
    \end{tabular}
    \label{tab:stat}
\end{table}

\begin{table*}[!t]
\caption{Results of novel and overall diagnosis prediction performance on the MIMIC-III dataset with 5\% training data. 
The best results are highlighted in \textbf{bold} while the second best are \underline{underlined}.}
\centering
\begin{tabular}{lcccccccc}
\toprule
Task                                     & \multicolumn{4}{c}{Novel}                                    & \multicolumn{4}{c}{Overall}                                     \\
\cmidrule(lr){1-9}
\multicolumn{1}{c}{\multirow{2}{*}{Metric}} & \multicolumn{2}{c}{Visit-Level} & \multicolumn{2}{c}{Code-Level} & \multicolumn{2}{c}{Visit-Level} & \multicolumn{2}{c}{Code-Level} \\
\cmidrule(lr){2-9}
\multicolumn{1}{c}{}                        & P@5           & P@10           & Acc@5         & Acc@10        & P@10           & P@20           & Acc@10         & Acc@20        \\
\midrule
LLaMA3.1   & 1.01$_{\pm0.09}$ & 2.22$_{\pm0.47}$ & 0.43$_{\pm0.19}$ & 2.19$_{\pm0.45}$ & 41.19$_{\pm2.11}$ & 42.06$_{\pm1.03}$ & 30.46$_{\pm0.84}$ & 39.88$_{\pm0.71}$ \\
Qwen3      & 0.65$_{\pm0.05}$ & 2.76$_{\pm0.51}$ & 0.68$_{\pm0.16}$ & 2.54$_{\pm0.28}$ & 47.03$_{\pm1.05}$ & 48.61$_{\pm1.12}$ & 32.73$_{\pm0.66}$ & 46.10$_{\pm1.02}$ \\
\midrule
RETAIN        & 10.14$_{\pm0.17}$ & 16.41$_{\pm0.24}$ & 6.95$_{\pm0.18}$ & 15.50$_{\pm0.21}$ & 38.69$_{\pm0.18}$ & 45.62$_{\pm0.26}$ & 27.68$_{\pm0.24}$ & 44.15$_{\pm0.23}$ \\
StageNet      & 12.91$_{\pm0.12}$ & 20.39$_{\pm0.23}$ & 8.76$_{\pm0.18}$ & 19.27$_{\pm0.23}$ & 35.70$_{\pm0.17}$ & 43.59$_{\pm0.24}$ & 25.71$_{\pm0.23}$ & 42.83$_{\pm0.19}$ \\
TRANS         & 12.14$_{\pm0.16}$ & 18.39$_{\pm0.21}$ & 8.17$_{\pm0.13}$ & 17.03$_{\pm0.17}$ & 38.60$_{\pm0.18}$ & 45.77$_{\pm0.23}$ & 27.56$_{\pm0.22}$ & 44.29$_{\pm0.23}$ \\
\midrule
CGL           & 14.24$_{\pm0.17}$ & 20.39$_{\pm0.18}$ & 9.64$_{\pm0.12}$ & 19.10$_{\pm0.22}$ & 38.98$_{\pm0.20}$ & 45.98$_{\pm0.23}$ & 28.06$_{\pm0.23}$ & 44.89$_{\pm0.28}$ \\
HiTANet       & 12.90$_{\pm0.15}$ & 19.61$_{\pm0.18}$ & 8.71$_{\pm0.13}$ & 18.79$_{\pm0.24}$ & 34.94$_{\pm0.18}$ & 43.88$_{\pm0.24}$ & 25.25$_{\pm0.20}$ & 43.10$_{\pm0.26}$ \\
BoxCare       & 11.57$_{\pm0.14}$ & 17.56$_{\pm0.17}$ & 7.80$_{\pm0.12}$ & 16.68$_{\pm0.25}$ & 39.91$_{\pm0.16}$ & 46.08$_{\pm0.25}$ & 28.53$_{\pm0.18}$ & 44.80$_{\pm0.20}$ \\
BoxLM         & 9.79$_{\pm0.17}$  & 16.51$_{\pm0.19}$ & 6.47$_{\pm0.15}$ & 15.31$_{\pm0.24}$ & 45.44$_{\pm0.23}$ & 51.46$_{\pm0.28}$ & 32.21$_{\pm0.27}$ & 49.21$_{\pm0.24}$ \\
\midrule
\method(Retain)   & \underline{21.32$_{\pm1.63}$} & \underline{23.59$_{\pm1.17}$} & \underline{14.26$_{\pm0.72}$} & \underline{21.83$_{\pm0.41}$} & \underline{49.66$_{\pm0.58}$} & \underline{52.40$_{\pm0.56}$} & \underline{35.11$_{\pm0.73}$} & \underline{50.10$_{\pm1.07}$} \\
\method(BoxLM)     & \textbf{21.54$_{\pm1.60}$} & \textbf{24.41$_{\pm1.16}$} & \textbf{14.48$_{\pm0.64}$} & \textbf{22.05$_{\pm0.57}$} & \textbf{51.73$_{\pm0.72}$} & \textbf{55.05$_{\pm0.95}$} & \textbf{36.44$_{\pm0.47}$} & \textbf{52.26$_{\pm0.74}$} \\
\bottomrule
\end{tabular}
\label{tab:prediction_eperformance_m3}
\end{table*}

\begin{table}[t]
\centering
\caption{Results of novel diagnosis prediction performance on the MIMIC-IV dataset with 5\% training data. 
The best results are highlighted in \textbf{bold} while the second best are \underline{underlined}.}
\begin{tabular}{p{1.6cm}cccc}
\toprule
\multirow{2}{*}{\textbf{Metric}} 
& \multicolumn{2}{c}{Visit-Level} 
& \multicolumn{2}{c}{Code-Level} \\
\cmidrule(lr){2-3} \cmidrule(lr){4-5}
& P@5 & P@10 & Acc@5 & Acc@10 \\
\hline
LLaMA3.1             & 5.29$_{\pm0.76}$   & 6.13$_{\pm0.66}$  & 4.46$_{\pm0.56}$  & 7.31$_{\pm0.70}$   \\
Qwen3                & 5.14$_{\pm0.46}$   & 6.36$_{\pm0.61}$  & 4.36$_{\pm0.28}$  & 7.53$_{\pm0.74}$   \\
\hline
RETAIN                  & 8.12$_{\pm0.06}$   & 14.03$_{\pm0.08}$  & 5.71$_{\pm0.11}$  & 12.14$_{\pm0.15}$   \\
StageNet                & 7.81$_{\pm0.06}$  & 13.10$_{\pm0.07}$  & 5.38$_{\pm0.10}$ & 11.20$_{\pm0.11}$   \\
TRANS                   & 9.03$_{\pm0.07}$  & 14.35$_{\pm0.11}$  & 6.31$_{\pm0.14}$ & 12.32$_{\pm0.13}$   \\
\hline
CGL                     & 8.02$_{\pm0.04}$  & 14.54$_{\pm0.05}$  & 5.08$_{\pm0.07}$ & 11.71$_{\pm0.09}$   \\
HiTANet                 & 7.43$_{\pm0.07}$  & 10.88$_{\pm0.07}$  & 5.09$_{\pm0.05}$ & 9.24$_{\pm0.10}$   \\
BoxCare                 & 8.09$_{\pm0.05}$  & 14.21$_{\pm0.08}$  & 5.21$_{\pm0.10}$ & 11.81$_{\pm0.10}$   \\
BoxLM                   & 8.35$_{\pm0.09}$   & 15.08$_{\pm0.07}$ & 5.57$_{\pm0.12}$  & 12.12$_{\pm0.11}$   \\
\hline
\method (R)            & \underline{11.87$_{\pm0.98}$}  & \underline{16.24$_{\pm0.85}$}  & \underline{8.82$_{\pm0.46}$}   & \underline{15.41$_{\pm0.72}$}   \\
\method (B)        & \textbf{12.17$_{\pm0.76}$}    & \textbf{16.88$_{\pm0.28}$}     & \textbf{8.84$_{\pm0.45}$}      & \textbf{15.44$_{\pm0.26}$}   \\
\bottomrule
\end{tabular}
\label{tab:prediction_eperformance_m4_new}
\end{table}

\subsection{Experiment Settings}
\subsubsection{Datasets and Evaluation Protocols}
We evaluate \method on two widely-used EHR datasets: MIMIC-III~\cite{johnson2016mimic} and MIMIC-IV~\cite{johnson2018mimic}, following standard benchmarks~\cite{tan2025boxlm, chen2024predictive}.
Each dataset is preprocessed by mapping all ICD-9/10 codes to CCS categories, resulting in 285 and 842 unique CCS codes in MIMIC-III and MIMIC-IV, respectively. The statistics are summarized in Table~\ref{tab:stat}. 

To reflect realistic diagnostic workflows, we focus on next-visit prediction using patients with at least 2 visits. Given historical visits, the model predicts CCS codes for the next one. We consider two evaluation settings: (1) \textit{overall prediction,} where all ground-truth CCS codes are evaluated; and (2) \textit{novel prediction,} where only codes not present in the patient’s history are considered, evaluating the model's ability to identify clinically emerging diagnoses.


We report both visit-level and code-level metrics using standard multi-label measures: Precision@k (P@k) and Accuracy@k (Acc@k). Details and metric formulas are included in Appendix~\ref{app:metrics}.

\subsubsection{Baselines}
We compare \method with a comprehensive set of baseline methods from three categories: (1) LLM-based methods: We include LLaMA3.1-8B~\cite{dubey2024llama} and Qwen3-8B~\cite{yang2025qwen3}, which perform diagnosis prediction via direct prompting over EHR inputs. These models are tested with the same prompt template for fair comparison;  (2) Sequential modeling methods: RETAIN~\cite{choi2016retain}, StageNet~\cite{gao2020stagenet}, and TRANS~\cite{chen2024predictive}, which encode visit sequences with RNNs, LSTM or Transformer mechanisms for diagnosis prediction, respectively; (3) Structure-aware methods: CGL~\cite{lu2021collaborative}, HiTANet~\cite{luo2020hitanet}, BoxCare~\cite{lv2024boxcare}, and BoxLM~\cite{tan2025boxlm}, which incorporate EHR graph or hierarchical ontology to enhance the representation of medical concepts.
The details of the baselines are provided in Appendix~\ref{app:baselines}.

\begin{table*}[!t]
\caption{Ablation study of \method components on MIMIC-III.}
\centering
\begin{tabular}{lcccc|cccc}
\toprule
Task                                     & \multicolumn{4}{c}{Novel}                                    & \multicolumn{4}{c}{Overall}                                     \\
\cmidrule(lr){1-9}
\multicolumn{1}{c}{\multirow{2}{*}{Metric}} & \multicolumn{2}{c}{Visit-Level} & \multicolumn{2}{c}{Code-Level} & \multicolumn{2}{c}{Visit-Level} & \multicolumn{2}{c}{Code-Level} \\

\cmidrule(lr){2-9}
\multicolumn{1}{c}{}                        & P@5           & P@10           & Acc@5         & Acc@10        & P@10           & P@20           & Acc@10         & Acc@20        \\
\midrule
Qwen3-8B (Base) & 0.65$_{\pm0.05}$ & 2.76$_{\pm0.51}$ & 0.68$_{\pm0.16}$ & 2.54$_{\pm0.28}$ & 47.03$_{\pm1.05}$ & 48.61$_{\pm1.12}$ & 32.73$_{\pm0.66}$ & 46.10$_{\pm1.02}$ \\
\ \  + Candidate    & 9.61$_{\pm0.42}$  & 12.11$_{\pm0.57}$  & 6.45$_{\pm0.43}$ & 11.04$_{\pm0.62}$ & 50.81$_{\pm0.82}$  & 52.50$_{\pm0.74}$  & 35.89$_{\pm0.79}$ & 49.05$_{\pm0.69}$\\
\ \ \ \  + Prioritization  & 10.71$_{\pm0.77}$  & 14.34$_{\pm0.53}$  & 7.53$_{\pm0.68}$  & 13.95$_{\pm1.03}$  & 52.61$_{\pm0.76}$  & 54.19$_{\pm0.78}$  & 37.17$_{\pm0.55}$ & 51.52$_{\pm0.70}$  \\
\ \ \ \ \ \  + Relational (Full)  & 21.54$_{\pm1.60}$  & 24.41$_{\pm1.16}$  & 14.48$_{\pm0.64}$ & 22.05$_{\pm0.57}$  & 51.73$_{\pm0.72}$  & 55.05$_{\pm0.95}$  & 36.44$_{\pm0.47}$ & 52.26$_{\pm0.74}$ \\

\bottomrule
\end{tabular}
\label{tab:ablation}
\end{table*}


\subsubsection{Implementation Details}
We follow the same experimental setup as used in TRANS~\cite{chen2024predictive} and BoxLM~\cite{tan2025boxlm}. Both datasets are split into training, validation, and test sets using a 7:1:2 ratio at patient level. All compared methods are trained using Adam optimizer, and their hyperparameters are tuned as recommended in their original papers. We set the embedding dimension to 16 for all models. For the in-context reasoning, we employ Qwen3-8B as the foundation model. All experiments are conducted with 5-fold cross-validation. 


\subsection{Main Results and Analysis (RQ1)}

To assess the overall effectiveness of \method in diagnosis prediction, we compare its performance against a range of baselines on both MIMIC-III and MIMIC-IV datasets under 5\% training settings (shown in Tables~\ref{tab:prediction_eperformance_m3}, ~\ref{tab:prediction_eperformance_m4_new} and~\ref{tab:prediction_eperformance_m4_all}). The performance comparison under varying ratios of training data scenarios (e.g., 1\%, 5\%, 10\%, 50\%, and 100\%) is provided in Appendix~\ref{appendix:train-size}. 

In general, \method consistently outperforms all baseline methods, with performance gains ranging from reasonably large (6.2\% achieved with Acc@20 on MIMIC-III) to significantly large (51.26\% achieved with P@5 on MIMIC-III). These gains answer RQ1, showing that our in-context reasoning strategy can leverage deep model guidance to enhance LLM prediction accuracy and interpretability.

\noindent\textbf{Performance on Novel Diagnosis Prediction.}
We observe the most pronounced improvements in the novel diagnosis setting, which evaluates the model's ability to identify CCS codes that have not appeared in a patient’s history. 
This task is particularly challenging for LLM-only approaches that tend to rely on shallow textual associations or the repetition of previously seen diagnoses. For example, Qwen3 achieves only 2.76 P@10 and 2.54 Acc@10 on MIMIC-III, while \method (BoxLM) improves these metrics to 24.41 and 22.05 respectively, achieving a relative improvement of more than 700\%. These results highlight the necessity of relational evidence construction via deep models.

Compared to deep learning baselines, \method also yields substantial gains across both temporal (e.g., RETAIN) and structure-aware models (e.g., BoxLM). On MIMIC-III, RETAIN achieves 16.41 P@10 and 15.50 Acc@10, while our \method (RETAIN) variant achieves 23.59 and 21.83, with relative improvements of 43.8\% and 40.8\%. Similarly, BoxLM's performance improves from 16.51 to 24.41 in P@10 and from 15.31 to 22.05 in Acc@10 when integrated into \method. These results indicate that our framework successfully enhances the generalization ability of deep models by shifting the LLM’s focus to novel but informative diagnoses.

\noindent\textbf{Performance on Overall Diagnosis Prediction.}
The overall diagnosis prediction results on MIMIC-III are presented in Table~\ref{tab:prediction_eperformance_m3}, while the corresponding results on MIMIC-IV are shown in Table~\ref{tab:prediction_eperformance_m4_all}, with the complete analysis provided in Appendix~\ref{appendix:overall}.

\method consistently outperforms both LLM-only and deep learning baselines. On MIMIC-III, \method (BoxLM) achieves 51.73 P@10 and 36.44 Acc@10, improving upon the BoxLM backbone by +6.29 and +4.23 respectively. Compared to Qwen3 (47.03 P@10, 32.73 Acc@10), our model yields notable absolute gains while maintaining interpretability. We further observe consistent gains on MIMIC-IV, where \method (Retain) achieves 55.42 P@10 and outperforms all baselines.
These improvements reflect the benefit of guiding LLMs with deep model-derived evidence and structured prompts.

\begin{figure}
    \centering
    \includegraphics[width=\linewidth]{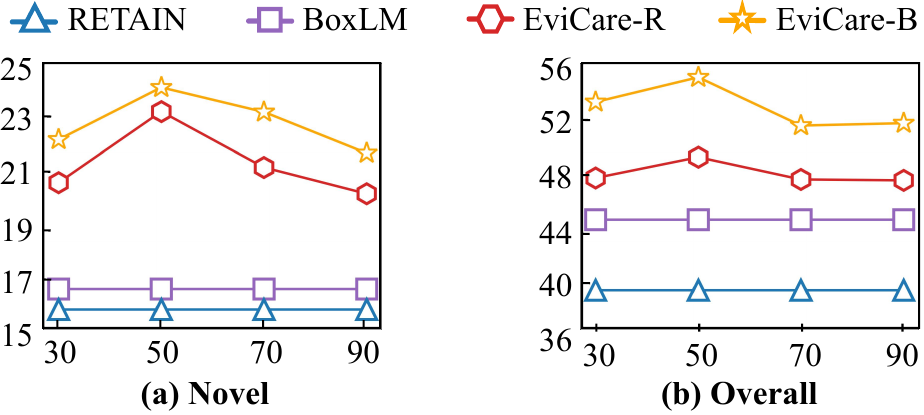}
    \caption{P@10 performance on novel and overall diagnosis prediction under varying candidate CCS sizes (MIMIC-III).}
    \label{fig:hyper_ccs_size}
\end{figure}

\subsection{Ablation Study (RQ2)}

To examine the contribution of each component in \method, we conduct ablation experiments on MIMIC-III, incrementally adding components: \textit{Candidate Selection (Candidate)}, \textit{Evidential Prioritization (Prioritization)}, and \textit{Relational Evidence (Relational)}. Table~\ref{tab:ablation} reports the results on both novel and overall diagnosis tasks. We observe general performance gains as more components are added.

The most substantial improvements appear in the novel setting. Starting from the base configuration, which achieves only 2.76 P@10 and 2.54 Acc@10, the inclusion of candidate selection boosts performance to 12.11 and 11.04, respectively. The results show that narrowing the reasoning space significantly improves precision. Adding prioritization further improves P@10 to 14.34 and Acc@10 to 13.95. Finally, relational brings the largest jump, reaching 24.41 P@10 and 22.05 Acc@10, demonstrating the critical role of structured evidence in uncovering clinically novel diagnoses.

While overall gains are more moderate, each component provides measurable improvements. 
Although overall P@10 slightly drops from 52.50  to 51.73 after adding symbolic relational evidence, novel prediction improves significantly (e.g., visit-level P@10 increase by 70.22\%). This trade-off reflects a shift toward broader clinical coverage and generalization, as top-20 metrics continue to improve despite minor top-10 fluctuations. For example, from the base to the full model, P@20 continuously increases from 48.61 to 55.05. 

\begin{table}[t]
\centering
\caption{Comparison with different reasoning paradigm.}
\label{tab:novel_reasoning}
\begin{tabular}{>{\centering\arraybackslash}m{0.95cm} >{\centering\arraybackslash}m{0.6cm} m{5.8cm}}
\toprule
\textbf{Method} & \textbf{P@10} & \textbf{Reasoning Example Highlights} \\
\midrule
\rowcolor{gray!20}
Qwen (CoT) & 3.17 & “\textit{Atrial flutter and lung edema} are often seen in \textit{heart failure} and \textit{pneumonia}” \\
Qwen (SC) & 4.16 & “\textit{Cardiomyopathies} and \textit{dysrhythmias} are strong indicators of heart failure” \\
\rowcolor{gray!20}
\method (Ours) & \textbf{24.58} & “AICD adjustment suggests \textit{conduction issues}, which links to nonhypertensive heart failure” \\
\bottomrule
\end{tabular}
\end{table}




\begin{table*}[]
\caption{Case study on a MIMIC-III patient comparing base model (Qwen3-8B), deep model (BoxLM) and \method under overall CCS prediction settings. Blue = Novel CCS in ground truth; Red = Incorrect prediction. Arrows (→) indicate evidence chains inferred from patient history to support predicted diagnoses.}
\begin{tabular}{p{0.12\linewidth} p{0.83\linewidth}}
\toprule
Method & Diagnosis Prediction Set \\
\hline
\multirow{4}*{\shortstack{ \textbf{Input EHR} \\ \textbf{Data}}} &  \textbf{ICD} :\textsf{Cardiac complications, not elsewhere classified (\textbf{CNC}); Acute pancreatitis (\textbf{AP}); Peritoneal adhesions postoperative postinfection (\textbf{Papp}); Congestive heart failure, unspecified (\textbf{CHFu}); ...} \newline
\textbf{CCS} : \textsf{Complications of surgical procedures or medical care (\textbf{CSPM}); Pancreatic disorders (not diabetes) (\textbf{PD}); Other gastrointestinal disorders (\textbf{OGD}); Congestive heart failure; nonhypertensive (\textbf{CHFn}); ...}\\
\hline
 \multirow{2}*{\shortstack{ \textbf{Ground-Truth} \\ Num:10}} &  \blue{Coronary atherosclerosis and other heart disease(\textbf{CAHD})}; \textbf{CHFn}, \blue{Peripheral and visceral atherosclerosis(\textbf{PVA});} 
 \textbf{CSPM}; 
 \blue{Deficiency and other anemia(\textbf{DOA}); Acute posthemorrhagic anemia(\textbf{APA}); Essential hypertension(\textbf{EH});...} \\
\hline
\multirow{6}*{\shortstack{\textbf{Qwen3-8B} \\ Correct: 3 }}      & \textbf{CHFn} (\textsf{...directly listed in the patient's history...is a complication of \textbf{AP}..can lead respiratory failure...also in history}.) \\
& \textbf{CSPM} (\textsf{...listed in the patient's history...linked to \textbf{Papp} and \textbf{AP}...}) \\
& \blue{\textbf{PVA}} (\textsf{...patient's history of heart disease and \textbf{CNC} supports this possibility...})\\
& \textcolor{red}{Pulmonary heart disease} (...patient history has \textbf{CHFn},...lead pulmonary hypertension an pulmonary heart disease.)\\
& \textcolor{red}{Nonspecific chest pain} (...a symptom of various cardiac and non-cardiac conditions.) ...\\
\hline
 \multirow{1}*{\shortstack{{\textbf{BoxLM} (3)}}} &   \textbf{CSPM}, \blue{\textbf{CAHD}}, \textbf{CHFn}, \textcolor{red}{Cardiac dysrhythmias, Septicemia (except in labor)} ...\\
\hline
 \multirow{6}*{\shortstack{\textbf{\method} \\  Correct: 5}}  &  \textbf{CSPM} (\textsf{...supported by history of \textbf{CNC} and \textbf{Papp},... -> \textbf{CSPM} }) \\
& \textbf{CHFn} (\textsf{...patient's history has \textbf{CHFn} and \textbf{CHFu}, ... -> \textbf{CHFn}...}) \\
& \blue{\textbf{DOA}} (\textsf{...has history of \textbf{Papp}...-> \textcolor{blue}{\textbf{DOA}}, \textbf{Papp} -> nutritional deficiencies and anemia.}) \\
& \blue{\textbf{EH}} (\textsf{...has history of \textbf{AP}...-> \textcolor{blue}{\textbf{EH}}, \textbf{AP} -> hypertension due to systemic inflammation...}) \\
& \blue{\textbf{CAHD}} (\textsf{...has history of Subendocardial infarction, initial episode of care -> \textcolor{blue}{\textbf{CAHD}},...due to the pathophysiology of myocardial infarction.}) ... \\
\bottomrule
\end{tabular}
\label{tab:case_study}
\end{table*}

\subsection{Comparison with Generic Reasoning (RQ3)}

We compare \method with two representative reasoning paradigms: Chain-of-Thought (CoT) and Self-Consistency (SC). All methods use Qwen3-8B as the base model and operate over the same candidate CCS set for fairness. As shown in Table~\ref{tab:novel_reasoning}, \method achieves a P@10 of 24.58 under the novel diagnosis setting, significantly outperforming Qwen (CoT) (3.17) and Qwen (SC) (4.16).

Although CoT encourages step-by-step reasoning and SC promotes output diversity through sampling, both approaches lack EHR-derived structure and clinical grounding. As a result, their reasoning often relies on superficial semantic correlations. For example, CoT may associate \textit{lung edema} with \textit{pneumonia}, leading to plausible yet clinically misaligned conclusions. SC, on the other hand, tends to produce generic statements (e.g., cardiomyopathies suggest heart failure), which offers little patient-specific insight.

In contrast, \method focuses on novel candidates and explicitly links them to the patient's history using symbolic relational evidence guided by deep model predictions. This strategy enables more targeted and clinically informed reasoning. Additional performance metrics and representative examples are provided in Appendix~\ref{appendix:reason}.

\subsection{Generalization Analysis (RQ4)}

To assess the robustness and generalizability of \method, we analyze its performance under varying candidate set sizes and across different LLM backbones, focusing on the MIMIC-III dataset.

\noindent\textbf{Candidate Size.} Figure~\ref{fig:hyper_ccs_size} illustrates the impact of varying candidate set sizes on P@10. We observe that performance on both diagnosis prediction peaks when $K=50$, after which performance slightly declines. This trend suggests that smaller candidate sets may omit true positives, while overly large sets introduce noise, weakening the focus of in-context reasoning. Thus, moderate pruning (e.g., $K=50$) best facilitates the LLM’s evidence-driven prediction process.

\begin{table}[t]
\centering
\caption{Generalization of \method across different LLMs on MIMIC-III. The metric used is Precision@10.}
\label{tab:llm_generalization}
\begin{tabular}{>{\centering\arraybackslash}m{1.4cm} >{\centering\arraybackslash}m{1.2cm} >{\centering\arraybackslash}m{1.2cm} >{\centering\arraybackslash}m{1.2cm} >{\centering\arraybackslash}m{1.4cm}}
\toprule
\textbf{LLM} & \textbf{Deep} & \textbf{Novel} & \textbf{Overall} & \textbf{Avg. Imp.} \\
\midrule
\multirow{3}{*}{\textbf{LLaMA3.1}} 
& None   & 2.22  & 41.19 & -- \\
& RETAIN & 20.73 & 48.21 & 425.4\% \\
& BoxLM  & 22.15 & 50.09 & 459.4\% \\
\midrule
\multirow{3}{*}{\textbf{Qwen3}} 
& None   & 2.76  & 47.03 & -- \\
& RETAIN & 23.59 & 49.66 & 380.1\% \\
& BoxLM  & 24.41 & 51.73 & 397.4\% \\
\bottomrule
\end{tabular}
\label{tab:llm_backbone}
\end{table}


\noindent\textbf{LLM Backbones.} To assess whether \method generalizes across different LLM backbones, we fix the underlying deep model (RETAIN or BoxLM) and compare the performance between two representative LLMs (LLaMA3.1 and Qwen3).
As shown in Table~\ref{tab:llm_backbone}, both LLMs achieve substantial gains over their standalone baselines, indicating that \method consistently improves reasoning capabilities across model types.

Notably, while Qwen3 shows stronger standalone performance (e.g., 2.76 vs. 2.22 P@10 on novel diagnoses), LLaMA3.1 combined with BoxLM (50.09) surpasses outperforms Qwen3 with RETAIN (49.66) in overall precision. This indicates that \method’s structured backbone design can compensate for weaker LLM capacity.
These trends demonstrate that \method generalizes well across diverse LLM architectures, and its design effectively activates reasoning regardless of the LLM backbone.

\subsection{Interpretable Case Study (RQ5)}

We present a case study to highlight how \method enhances both accuracy and interpretability in complex diagnostic scenarios. Compared to Qwen3-8B and BoxLM, \method better captures clinically relevant yet previously unseen diagnoses by integrating deep model predictions with symbolic co-occurrence evidence.

As shown in Table~\ref{tab:case_study}, Qwen3-8B correctly predicts 3 CCS codes, mainly relying on history repetition. Its explanation often overgeneralizes, such as linking \textit{Pulmonary heart disease} to a vague mention of heart failure, despite no evidence of pulmonary hypertension. It also exhibits hallucinations, e.g., \textit{Nonspecific chest pain}.
Although BoxLM achieves similar accuracy with two historical and one novel diagnosis, it fails to provide explanation, limiting its clinical interpretability.

In contrast, \method correctly predicts 5 ground-truth CCS codes, including 3 novel ones, and provides clear, clinically grounded explanations. For example,
\textit{Essential hypertension} is associated with \textit{acute pancreatitis} based on known inflammatory mechanisms; \textit{Deficiency and other anemia} is explained by the presence of \textit{peritoneal adhesions (postinfection)}. This demonstrates \method’s ability to move beyond surface-level pattern matching toward evidence-guided, interpretable reasoning.

%% file: chapter/conclusion.tex
\section{Conclusion}

In this paper, we propose \method, a deep model-guided in-context reasoning framework for LLM-based diagnosis prediction.
By incorporating candidate selection, evidential prioritization, and relational evidence construction, \method enables LLMs to make targeted and interpretable predictions over complex clinical histories.
Experiments on MIMIC-III and MIMIC-IV show that \method consistently outperforms both LLM-only and deep model-only competitors, particularly in predicting novel diagnoses that are unseen previously but clinically important.
This task is clinically valuable for discovering emerging pathological patterns and comorbidities. Extensions to other applications, such as drug recommendation, will be explored in future work.


%% file: chapter/appendix.tex
\section*{APPENDIX}

\section{Detail of Evaluation Metrics}
\label{app:metrics}

We evaluate diagnosis prediction at visit and code levels using Precision@k (P@k) and Accuracy@k (Acc@k).

\paragraph{Visit-Level Precision@k.}  
This metric computes the proportion of correctly predicted codes within the top-$k$ predictions for each individual visit. Formally, for a visit $t$, let $\hat{Y}_t$ be the top-$k$ predicted codes, and $Y_t$ the ground-truth codes for the next visit. The visit-level precision is defined as:
\begin{equation}
\text{P@k}_{\text{visit}} = \frac{1}{|T|} \sum_{t=1}^{|T|} \frac{|\hat{Y}_t \cap Y_t|}{\min(k, |Y_t|)},
\end{equation}
where $T$ is the set of all visits in the test set. The normalization ensures fairness when $|Y_t| < k$.

\paragraph{Code-Level Accuracy@k.}  
This metric measures the overall proportion of correctly predicted codes across all patients, weighted by total ground-truth codes. Let $P$ denote the set of test patients, and for each patient $t$, $Y_t$ and $\hat{Y}_t$ are defined similarly. The code-level accuracy is given by:
\begin{equation}
\text{Acc@k}_{\text{code}} = \frac{\sum_{t \in P} |\hat{Y}_t \cap Y_t|}{\sum_{t \in P} |Y_t|}.
\end{equation}

\paragraph{Novel Diagnosis Evaluation.}  
For the \textit{novel diagnosis prediction} setting, a predicted CCS code is counted as correct only if it satisfies:
\begin{itemize}
  \item[(i)] It appears in the ground-truth codes $Y_t$, and
  \item[(ii)] It does not appear in any of the patient’s previous visits.
\end{itemize}
This stricter evaluation reflects the model’s ability to identify unseen or emerging diagnoses.

\section{Baseline Details}
\label{app:baselines}

We provide a summary of all baseline models used in our experiments. The models are organized into four categories based on their core design characteristics.
\subsection{LLM-based Methods}
\begin{itemize}[leftmargin=10pt]
    \item \textbf{LLaMA3.1-8B}~\cite{dubey2024llama}: An open-source LLM for efficient language understanding and generation.
    \item \textbf{Qwen3-8B}~\cite{yang2025qwen3}: supports multilingual instruction following and complex reasoning with strong alignment performance.
    \item \textbf{Chain-of-Thought (CoT)}~\cite{wei2022chain}: prompts LLMs to generate intermediate reasoning steps for better decision-making.
    \item \textbf{Self-Consistency (SC)}~\cite{wang2022self}: selects the most consistent answer from multiple reasoning paths to enhance reliability.
\end{itemize}

\subsection{Sequential Modeling Methods}
\begin{itemize}[leftmargin=10pt]
    \item \textbf{RETAIN}~\cite{choi2016retain}: applies RNNs with a reverse-time attention mechanism to predict patient diagnoses.
    \item \textbf{StageNet}~\cite{gao2020stagenet}: models disease progression using stage-aware LSTMs and stage-adaptive convolutions.
    \item \textbf{TRANS}~\cite{chen2024predictive}: constructs temporal heterogeneous graphs to capture dynamics and structure in EHRs.
\end{itemize}


\subsection{Ontology-aware Methods}
\begin{itemize}[leftmargin=10pt]
    \item \textbf{CGL}~\cite{lu2021collaborative}: models patient-disease interactions and external knowledge via collaborative graph learning.
    \item \textbf{HiTANet}~\cite{luo2020hitanet}: models temporal patterns in EHRs using a hierarchical time-aware transformer.
    \item \textbf{BoxCare}~\cite{lv2024boxcare}: represents inclusion and exclusion relations among diseases via box embeddings in a structured space.
    \item \textbf{BoxLM}~\cite{tan2025boxlm}: aligns medical concepts through unified structure- and semantics-aware box embeddings.
\end{itemize}

\section{Ontology Graph Construction and Data Sources}
\label{sec:source}
We construct the ontology graph by mapping ICD-10-CM and ICD-10-PCS codes to multi-level clinical categories defined in the Clinical Classifications Software Refined (CCSR)\footnote{https://hcup-us.ahrq.gov/toolssoftware/ccsr/ccs\_refined.jsp}.

\section{Full Overall Diagnosis Prediction Analysis}
\label{appendix:overall}
In the overall diagnosis setting, which evaluates a model’s ability to predict all ground-truth CCS codes. \method achieves consistent improvements over both LLM-only and deep learning baselines.

LLM-only models such as Qwen3 and LLaMA3.1 demonstrate strong performance due to their memorization of historical patterns and semantic associations in clinical texts. For instance, Qwen3 achieves 50.99 P@20 and 46.06 Acc@20 on MIMIC-IV, outperforming several traditional deep models (e.g., RETAIN and StageNet), which primarily rely on sequential visit modeling. However, their predictions are often biased toward frequent diagnoses and lack fine-grained reasoning over complex patient histories. By contrast, \method supplements LLMs with candidate selection and evidential scaffolding derived from deep models, enabling more accurate and context-aware reasoning with improvement.

\begin{table}[h]
\caption{Results of overall diagnosis prediction performance on the MIMIC-IV dataset with 5\% training data. 
The best results are highlighted in \textbf{bold} while the second best are \underline{underlined}.}
\begin{tabular}{p{1.55cm}cccc}
\toprule
\multirow{2}{*}{\textbf{Metric}} 
& \multicolumn{2}{c}{Visit-Level} 
& \multicolumn{2}{c}{Code-Level} \\
\cmidrule(lr){2-3} \cmidrule(lr){4-5}
& P@10 & P@20 & Acc@10 & Acc@20 \\
\hline
LLaMA3.1             & 38.32$_{\pm1.02}$   & 43.62$_{\pm1.36}$  & 31.22$_{\pm0.63}$  & 33.82$_{\pm0.72}$   \\
Qwen3                & 50.42$_{\pm0.90}$   & 50.99$_{\pm1.20}$  & 36.79$_{\pm0.69}$  & 46.06$_{\pm0.78}$   \\
\hline
RETAIN                  & 43.19$_{\pm0.08}$   & 49.05$_{\pm0.07}$  & 31.49$_{\pm0.07}$  & 45.60$_{\pm0.09}$   \\
StageNet                & 37.69$_{\pm0.08}$   & 43.45$_{\pm0.07}$  & 27.69$_{\pm0.06}$  & 40.89$_{\pm0.07}$   \\
TRANS                   & 36.00$_{\pm0.09}$   & 41.95$_{\pm0.08}$  & 26.52$_{\pm0.07}$  & 39.83$_{\pm0.08}$   \\
\hline
CGL                     & 32.48$_{\pm0.05}$   & 38.01$_{\pm0.07}$  & 22.42$_{\pm0.06}$  & 36.72$_{\pm0.10}$   \\
HiTANet                 & 27.34$_{\pm0.07}$   & 34.00$_{\pm0.11}$  & 20.07$_{\pm0.09}$  & 32.54$_{\pm0.12}$   \\
BoxCare                 & 34.61$_{\pm0.09}$   & 40.33$_{\pm0.12}$  & 24.29$_{\pm0.08}$  & 36.50$_{\pm0.10}$   \\
BoxLM                   & 43.02$_{\pm0.07}$   & 48.12$_{\pm0.08}$  & 31.20$_{\pm0.06}$  & 44.44$_{\pm0.09}$   \\
\hline
\method (R)            & \textbf{55.42$_{\pm0.74}$}  & \underline{57.12$_{\pm0.61}$}  & \textbf{40.41$_{\pm0.51}$}   & \underline{53.27$_{\pm0.65}$}   \\
\method (B)            & \underline{55.04$_{\pm0.76}$}    & \textbf{57.63$_{\pm0.61}$}     & \underline{40.17$_{\pm0.65}$}      & \textbf{53.80$_{\pm0.67}$}   \\
\bottomrule
\end{tabular}
\label{tab:prediction_eperformance_m4_all}
\end{table}

In contrast, deep models like BoxLM offer strong structural priors by encoding hierarchical and co-occurrence-based relationships between diagnoses, but they may lack interpretability and flexibility in complex clinical scenarios. \method addresses these limitations by using the deep model to generate a semantically meaningful candidate space and relational evidence, which are then synthesized through LLM-based reasoning. For instance, \method (BoxLM) improves upon its backbone by achieving 57.63 P@20 and 53.80 Acc@20, marking absolute gains of +9.51 and +9.36, respectively. This demonstrates that our hybrid design effectively combines the structural precision of deep models with the semantic generalization of LLMs. These benefits generalize well to MIMIC-IV, further validating the scalability and robustness of our framework across datasets with varying label spaces and patient distributions.

\section{Impact of Training Size}
\label{appendix:train-size}

\begin{figure}[h]    
\centering
    \includegraphics[width=1\linewidth]{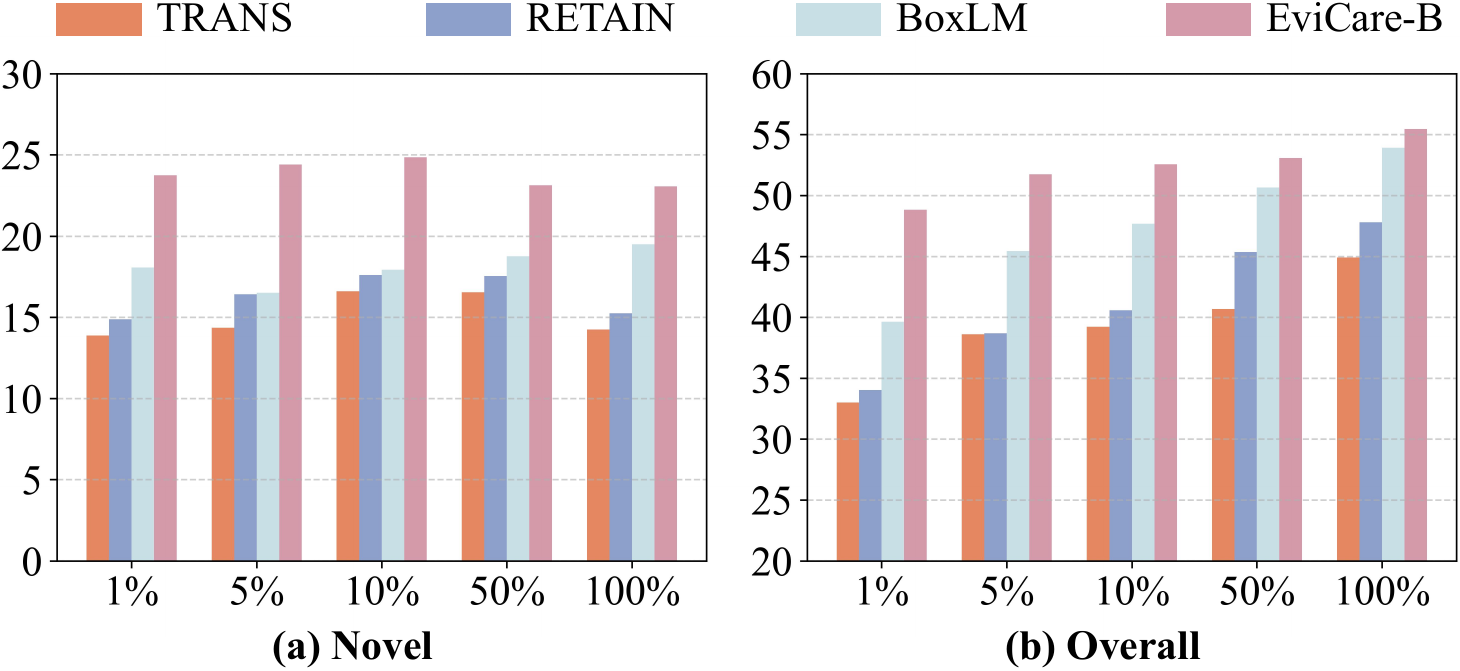}
    \caption{P@10 performance on novel and overall diagnosis prediction under different training data ratios (MIMIC-III).}
    \label{fig: train_size_ratio}
\end{figure}

Figure~\ref{fig: train_size_ratio} reports P@10 performance of \method and three base models (TRANS, RETAIN, BoxLM) under varying training data ratios (1\%, 5\%, 10\%, 50\%, 100\%) on MIMIC-III. Across all settings, \textsc{EviCare-B} consistently achieves higher accuracy on both novel and overall diagnosis tasks. Even with only 1\% of training data, \textsc{EviCare-B} outperforms base models trained on significantly more data, highlighting its robustness under limited supervision.


\section{Full Reasoning Strategy Study}
\label{appendix:reason}

\begin{table}[h]
\centering
\caption{Performance Comparison of Different Reasoning Strategies.}
\begin{tabular}{l|cccc}
\toprule
\textbf{Novel} & P@5 & P@10 & Acc@5 & Acc@10 \\
\midrule
Qwen (CoT)         & 1.25 & 3.17 & 1.63 & 3.07 \\
Qwen (SC)          & 2.09 & 4.26 & 2.71 & 4.02 \\
\method       & 22.03 & 24.58 & 14.56 & 22.12 \\
\midrule
\textbf{Overall} & P@10 & P@20 & Acc@10 & Acc@20 \\
\midrule
Qwen (CoT)         & 41.91 & 45.66 & 29.12 & 43.80 \\
Qwen (SC)          & 42.23 & 46.09 & 30.15 & 44.26 \\
\method       & 52.01 & 55.23 & 36.67 & 52.11 \\
\bottomrule
\end{tabular}
\label{tab:llm_reasoning}
\end{table}

Table~\ref{tab:llm_reasoning} shows the prediction performance under different reasoning strategies. Table~\ref{tab:case_single} presents an example of predicting a novel diagnosis \textit{Congestive heart failure; nonhypertensive} using different reasoning methods. Although the target condition was correctly predicted by all methods, the generated explanations showed differences.

We retain only the clinically relevant entries from the patient's historical EHR for clarity. All models were given the same candidate set and input. Among the three structured prompting (SC) trials, two generated plausible explanations, while one failed to associate the history with the target diagnosis.

In contrast, our method (EviCare) generates a more comprehensive explanation based on relational evidence. Specifically, it links multiple historical conditions—\textit{Cardiac dysrhythmias}, \textit{Other primary cardiomyopathies}, and the \textit{Fitting and adjustment of an automatic implantable cardiac defibrillator (AICD)}—into a coherent causal chain.

This explanation is clinically supported. The implantation of an AICD is commonly indicated in patients with severe conduction disorders or ventricular arrhythmias, which are well-known precursors of sudden cardiac death and nonhypertensive heart failure. Such conduction abnormalities impair cardiac synchronization and function, directly contributing to heart failure.

Thus, the path from AICD adjustment, through conduction disorders, to nonhypertensive heart failure, reflects a medically established progression supported by evidence from clinical trials such as MADIT-II and SCD-HeFT. This validates the relational evidence used by our method and illustrates its advantage in generating faithful, interpretable justifications for novel diagnosis prediction.

\begin{table}[h]
\centering
\small
\caption{Explanation comparison for a novel diagnosis prediction case.}
\label{tab:case_single}
\begin{tabular}{p{2.1cm}p{5.9cm}}
\toprule
\textbf{Patient History} & Other primary cardiomyopathies; Paroxysmal ventricular tachycardia; Fitting and adjustment of AICD; Cardiac dysrhythmias; Conduction disorders; ... \\
\hline
\textbf{Target Diagnosis} & Congestive heart failure; nonhypertensive \\
\midrule
\textbf{Qwen (CoT)} & The patient has a history of "Other primary cardiomyopathies" and "Paroxysmal ventricular tachycardia" which are both associated with heart failure. \\
\midrule
\multirow{3}{*}{\textbf{Qwen (SC)}} 
& 1. The patient has "Other primary cardiomyopathies" and "Cardiac dysrhythmias", both of which are precursors or components of nonhypertensive heart failure. Although not explicitly stated, the progression is clinically plausible. \\
& 2. The patient’s history includes "Other primary cardiomyopathies" (ICD) and "Cardiac dysrhythmias" (CCS), which are strong indicators of heart failure. Cardiomyopathies can progress to heart failure, and dysrhythmias like paroxysmal ventricular tachycardia are common in heart failure. \\
& 3. (No relevant prediction) \\
\midrule
\textbf{EviCare (Ours)} & This is supported by the patient's history of cardiac dysrhythmias, other primary cardiomyopathies, and AICD adjustment. These conditions are often associated with heart failure, especially in the context of chronic disease and device management. \\
\hline
\textbf{Relational  \newline  Evidence} & {"Fitting and adjustment of automatic implantable cardiac defibrillator"} → {"Conduction disorders"} → {"Congestive heart failure; nonhypertensive"} \\
\bottomrule
\end{tabular}
\end{table}